\documentclass[letterpaper, 10 pt, journal, twoside]{IEEEtran}
\usepackage{amsmath,amsfonts}
\usepackage{algorithmic}
\usepackage{algorithm}
\usepackage{array}
\usepackage[caption=false,font=normalsize,labelfont=sf,textfont=sf]{subfig}
\usepackage{textcomp}
\usepackage{stfloats}
\usepackage{url}
\usepackage{verbatim}
\usepackage{graphicx}
\usepackage{cite}
\usepackage{makecell}
\usepackage{booktabs}
\usepackage{arydshln} 
\usepackage{multirow}
\usepackage{balance}
\usepackage{fancyhdr}

\usepackage{amssymb}
\usepackage{color}
\usepackage{array}

\usepackage[switch]{lineno}

\usepackage{tikz,xcolor,hyperref}
\definecolor{lime}{HTML}{A6CE39}
\DeclareRobustCommand{\orcidicon}{%
    \begin{tikzpicture}
    \draw[lime, fill=lime] (0,0) 
    circle [radius=0.16] 
    node[white] {
   {\fontfamily{qag}\selectfont \tiny ID}};    \draw[white, fill=white] (-0.0625,0.095) 
    circle [radius=0.007];    \end{tikzpicture}
    \hspace{-2mm}}
\foreach \x in {A, ..., Z}{%
    \expandafter\xdef\csname orcid\x\endcsname{\noexpand\href{https://orcid.org/\csname orcidauthor\x\endcsname}{\noexpand\orcidicon}}
    }

\hypersetup{hidelinks,
	colorlinks=true,
	allcolors=blue,
	pdfstartview=Fit,
	breaklinks=true}

\begin{document}


\IEEEpubidadjcol
\title{HybridOcc: NeRF Enhanced Transformer-based Multi-Camera 3D Occupancy Prediction}

\author{Xiao Zhao\orcidA{}, Bo Chen, Mingyang Sun, Dingkang Yang, Youxing Wang, Xukun Zhang, Mingcheng Li, Dongliang Kou, Xiaoyi Wei, and Lihua Zhang\orcidF{},~\IEEEmembership{Member,~IEEE}
\thanks{Manuscript received: February 4, 2024; Revised: April 8, 2024; Accepted: June 7, 2024. This paper was recommended for publication by Editor Cesar Cadena Lerma upon evaluation of the Associate Editor and Reviewers’ comments. This work was supported in part by the National Key R\&D Program of China under No.2021ZD0113503, Shanghai Municipal Science and Technology Major Project under No.2021SHZDZX0103. \textit{(Corresponding author: Lihua Zhang.)}

Xiao Zhao, Mingyang Sun, Dingkang Yang, Youxing Wang, Xukun Zhang, Mingcheng Li, Dongliang Kou, and Xiaoyi Wei are with the Academy for Engineering and Technology, Fudan University, Shanghai 200000 China (e-mail: zhaox21@m.fudan.edu.cn)

Bo Chen and Youxing Wang are with the China FAW Group Corp., Ltd., Nanjing 211100, China.

Lihua Zhang is with the Academy for Engineering and Technology, Fudan University, Shanghai 200000 China, the Engineering Research Center of AI and Robotics, Ministry of Education, Shanghai, 200000 China, and also with the Jilin Provincial Key Laboratory of Intelligence Science and Engineering, Changchun, 130000, China (e-mail: lihuazhang@fudan.edu.cn).
}
\thanks{Digital Object Identifier (DOI): see top of this page.}}

\markboth{IEEE ROBOTICS AND AUTOMATION LETTERS. PREPRINT VERSION. ACCEPTED JUNE, 2024}
{Zhao \MakeLowercase{\textit{et al.}}: HybridOcc: NeRF Enhanced Transformer-based Multi-Camera 3D Occupancy Prediction}

\IEEEpubid{\begin{minipage}{\textwidth}\ \\[10pt] \centering
  0000-0000 \copyright 2024 IEEE. Personal use is permitted, but republication/redistribution requires IEEE permission.\\
  See http://www.ieee.org/publications standards/publications/rights/index.html for more information.
\end{minipage}}

\maketitle

\begin{abstract}
Vision-based 3D semantic scene completion (SSC) describes autonomous driving scenes through 3D volume representations. However, the occlusion of invisible voxels by scene surfaces poses challenges to current SSC methods in hallucinating refined 3D geometry. This paper proposes HybridOcc, a hybrid 3D volume query proposal method generated by Transformer framework and NeRF representation and refined in a coarse-to-fine SSC prediction framework. HybridOcc aggregates contextual features through the Transformer paradigm based on hybrid query proposals while combining it with NeRF representation to obtain depth supervision. The Transformer branch contains multiple scales and uses spatial cross-attention for 2D to 3D transformation. The newly designed NeRF branch implicitly infers scene occupancy through volume rendering, including visible and invisible voxels, and explicitly captures scene depth rather than generating RGB color. Furthermore, we present an innovative occupancy-aware ray sampling method to orient the SSC task instead of focusing on the scene surface, further improving the overall performance. Extensive experiments on nuScenes and SemanticKITTI datasets demonstrate the effectiveness of our HybridOcc on the SSC task.
\end{abstract}


\section{INTRODUCTION}
\IEEEPARstart{C}{amera-based} 3D scene understanding is a crucial component of the autonomous driving perception system. It involves acquiring accurate and comprehensive real-world 3D information, even with slight movement of the vehicle. In recent years, multi-camera systems have produced competitive results with Lidar in tasks such as depth estimation \cite{guizilini2022full,wei2023surroundocc} and 3D object detection \cite{philion2020lift,li2022bevformer}. Semantic scene completion (SSC) \cite{cao2022monoscene,wei2023surroundocc,Zhang2023OccFormerDT,li2023voxformer,tian2024occ3d,li2023fb} has recently gained more attention than 3D object detection. SSC is more appropriate for autonomous driving downstream tasks as it can represent scenes of arbitrary shapes and categories. However, inferring the comprehensive semantic scene from limited observation views is challenging.

\begin{figure}[t]
  \centering
   \centerline{\includegraphics[width=0.9\linewidth]{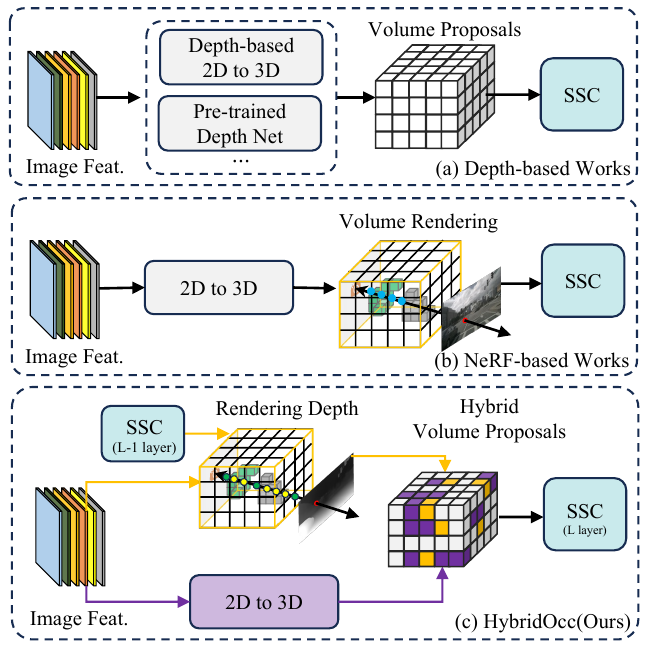}}
   \caption{Comparison of depth-based methods, NeRF-based and our HybridOcc. (a) Advanced methods such as FB-Occ \cite{li2023fb} and VoxFormer \cite{li2023voxformer} require additional depth prediction networks for generating 3D voxels. (b) NeRF-based methods \cite{zhang2023occnerf,pan2023renderocc} only focus on the visible surface (blue sampling points) of the scene and render based on the transformed 3D voxel features. (c) In HybridOcc, the NeRF branch combined with the Transformer branch gradually refines SSC from coarse-to-fine. We propose a 3D occupancy-aware ray sampling (yellow sampling points) to enable the model to focus on occupied voxels of all scenes rather than visible surfaces.}
   \label{fig1}
\end{figure}

\IEEEpubidadjcol

\begin{figure}[t]
  \centering
   \centerline{\includegraphics[width=1.0\linewidth]{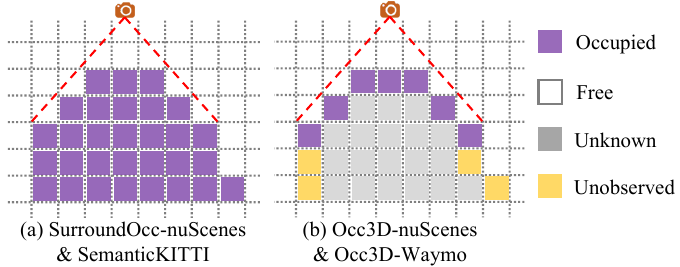}}
   \caption{Illustration of the 3D occupancy prediction data set. SurroundOcc-nuScenes \cite{wei2023surroundocc} and SemanticKITT \cite{behley2019semantickitti} evaluate all occupied voxels, while Occ3D-nuScenes and Occ3D-Waymo \cite{tian2024occ3d} only evaluate visible surfaces.}
   \label{figoccupancy prediction_compa}
\end{figure}

MonoScene \cite{cao2022monoscene} proposed directly lifting 2D images to 3D voxels through feature projection for the SSC task. Recently, some works \cite{huang2023tri,wei2023surroundocc,tian2024occ3d} proposed to lift multi-view camera features to 3D representation based on spatial cross-attention \cite{li2022bevformer}. In the coarse-to-fine framework proposed by Occ3D \cite{tian2024occ3d}, the performance is limited by the lack of depth signals. Other studies \cite{li2023voxformer,li2023fb,Zhang2023OccFormerDT} adopted additional depth estimation modules to improve the quality of 3D voxel representation, as shown in Fig. \ref{fig1}(a).
FB-Occ \cite{li2023fb} used a pre-trained depth prediction model and a depth-aware back-projection model to assist in generating 3D voxel features. However, most depth-based methods focus on the visible surface of the scene and lack inference of occluded regions. VoxFormer \cite{li2023voxformer} proposed an additional masked autoencoder-based module \cite{he2022masked} to consider the occluded voxels, but its cumbersome two-stage structure is not conducive to end-to-end model training. Various current methods have shown the importance of depth signals to the SSC task. Notably, there are currently two types of 3D occupancy datasets for autonomous driving with different functions. One is only evaluating the visible surface (Fig. \ref{figoccupancy prediction_compa}(b))\cite{tian2024occ3d}, while the other is for the complete occupation of the scene, that is, the SSC task (Fig. \ref{figoccupancy prediction_compa}(a))\cite{behley2019semantickitti,wei2023surroundocc}. This paper focuses more on the SSC task, which takes into account occluded objects or regions. Current SSC works \cite{li2023voxformer,wei2023surroundocc,Zhang2023OccFormerDT} mostly suffer from occlusion, making per-voxel features contain many ambiguities. Consequently, the occupancy prediction of occluded voxels still faces challenges.

The introduction of neural radiance fields (NeRFs) \cite{mildenhall2021nerf,cao2023scenerf} greatly improved the 3D scene reconstruction performance. SceneRF \cite{cao2023scenerf} designed a probabilistic ray sampling method for radiance field and applied it to the 3D reconstruction of autonomous driving scenes. Recently, some methods \cite{zhang2023occnerf,pan2023renderocc,huang2023selfocc} utilized the lifted 3D voxel features for depth and color rendering. Since NeRF-based 3D reconstruction methods focus on the visible surface of the scene, as shown in Fig. \ref{fig1}(b), the SSC task requires extra attention to voxel features in invisible regions. Therefore, a rough and direct application of the NeRF model on the SSC task may not be conducive to optimizing implicit function and the SSC task.

To address these challenges, we propose HybridOcc, a multi-camera semantic scene completion method. HybridOcc refines hybrid occupancy proposals generated by NeRF representation and Transformer architecture in a coarse-to-fine structure. As shown in Fig. \ref{fig1}(c), HybridOcc contains two branches. The Transforme branch, inspired by SurroundOcc \cite{wei2023surroundocc} and Occ3D \cite{tian2024occ3d}, uses learnable cross-attention to lift 2D images to 3D volume and gradually refine 3D volume queries from a coarse-to-fine structure. The NeRF branch innovatively adapts volume rendering with depth supervision to predict complete occupancy. Due to the NeRF optimization challenges posed by occlusions in autonomous driving scenes, we propose occupancy-aware ray sampling to optimize large radiance volumes. The implicit function is trained to serve the SSC task by taking occupancy-aware sampling points across visible and invisible voxels along the ray. The occupancy priors for each layer need to be carefully considered in the coarse-to-fine structure. Improved NeRF can hallucinate the occupancy of occluded invisible regions. The binary occupancy predicted by NeRF and coarse-grained Transformer is hybridized as a new volume query set to refine the semantic occupancy. In summary, our contributions are threefold:

1) We propose a novel complementary combination of contextual feature aggregation of the Transformer and depth supervision of NeRF. The hybrid occupancy proposals generated by NeRF representation and Transformer framework are refined end-to-end in a coarse-to-fine framework.

2) We introduce a novel depth-supervised neural radiance field that takes into account all visible and occluded invisible voxels for the SSC task. It adds depth signals to the coarse-to-fine SSC prediction framework and includes an occupancy-aware ray sampling strategy.

3) Extensive experiments demonstrate the effectiveness of our HybridOcc, which outperforms methods based on depth prediction networks such as FB-Occ and VoxFormer.

\section{RELATED WORKS}
\subsection{3D Semantic Scene Completion.}
3D semantic scene completion can provide a more detailed understanding of autonomous driving scenes. Some previous works \cite{song2017semantic,cai2021semantic}are studied in small-scale indoor scenes. With the release of the SemanticKITTI dataset \cite{behley2019semantickitti} and nuScenes dataset \cite{caesar2020nuscenes}, the SSC benchmark \cite{li2023sscbench,ssc4d,cao2024pasco} for large-scale autonomous driving scenes has been rapidly proposed recently. SurroundOcc \cite{wei2023surroundocc} and Occ3D \cite{tian2024occ3d} constructs nuScenes-based 3D occupancy prediction datasets respectively, one is oriented to the dense SSC task, and the other only evaluates occupancy of visible surfaces. These occupancy methods can be simply categorized into building 3D voxel features based on depth prediction \cite{li2023voxformer,Zhang2023OccFormerDT,li2023fb} and using Transformer-based learnable voxel feature aggregation \cite{huang2023tri,wei2023surroundocc,tian2024occ3d,ming2024occfusion}. Some methods \cite{Sparseocc,li2023fb,shi2024occfiner} introduce historical frame data to solve depth prediction and occlusion problems. OccFiner \cite{shi2024occfiner} proposes to implicitly capture and process multiple local frames. Additionally, some methods \cite{huang2023selfocc,zhang2023occnerf,pan2023renderocc} use NeRF \cite{mildenhall2021nerf} representation to explore occupancy task, but they focus more on reconstruction rather than SSC. We propose a method that combines the advantages of the Transformer paradigm and NeRF representation to enhance the SSC task performance.

\subsection{3D Scene Reconstruction.}
3D scene reconstruction aims to model 3D surface information from single- or multi-view 2D images. Early reconstruction methods focused on explicit representation of voxels \cite{xie2020pix2vox++} but now neural radiance fields (NeRF) \cite{mildenhall2021nerf,yang2023emernerf} and 3D Gaussian Splatting \cite{li2024memorize,cheng2024gaussianpro} are becoming more popular for implicit reconstruction. Considering that NeRF has the problem of slow rendering, some methods \cite{muller2022instant,hu2022efficientnerf,barron2023zip} improve the rendering speed while maintaining the rendering quality. Implicit reconstruction works \cite{Sun2021NeuralReconRC,Bovzivc2021TransformerFusionMR} based on image features extends object-level reconstruction to indoor scenes and is committed to building a generalized implicit network. \cite{Sun2021NeuralReconRC} and \cite{Bovzivc2021TransformerFusionMR} adopt a coarse-to-fine approach fuse multi-scale features to obtain more accurate 3D reconstruction of indoor scenes. SceneRF\cite{cao2023scenerf} proposes spherical U-Net and probabilistic ray sampling to expand NeRF for large-scale outdoor scenes. It is worth noting that 3D reconstruction under the NeRF paradigm requires the sampling points along the ray to be concentrated near the 3D surface for better rendering of color or semantics. However, for the SSC task, it makes more sense to concentrate the radiation field on occupied voxels.

\begin{figure*}[htb]
\centering
\centerline{\includegraphics[width=0.8\textwidth]{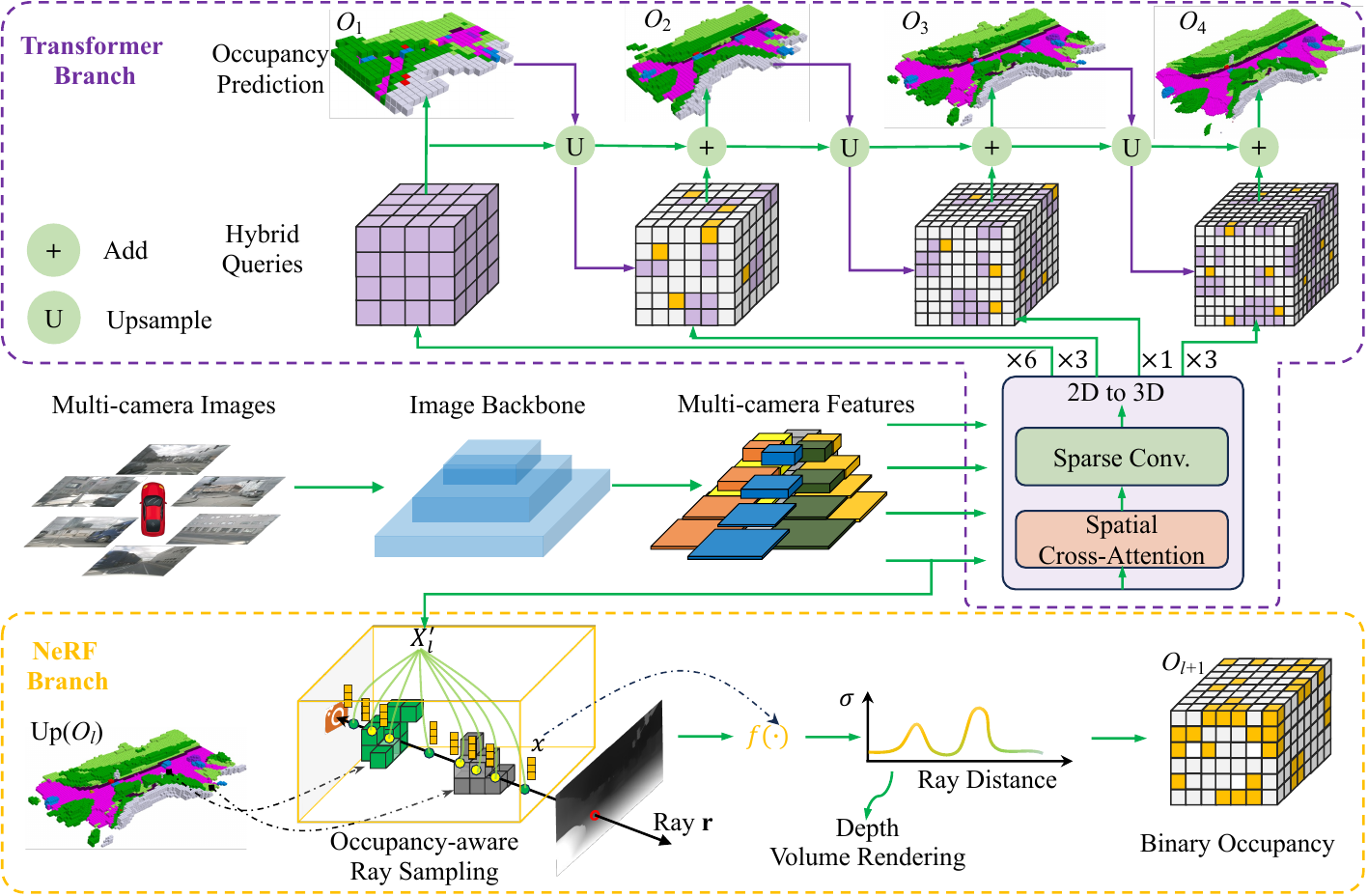}}
\caption{The pipeline of the proposed HybridOcc for multi-camera 3D semantic occupancy prediction. It consists of the image backbone for extracting multi-scale features and the dual branch composed of Transformer and NeRF to learn a sparse 3D feature volume from coarse to fine. The Transformer branch contains a 2D to 3D transformation module for lifting the 2D features to 3D volumes, and the NeRF branch obtains supervision from depth signals to enhance the Transformer branch.}
\label{Fig.Overall pipline}
\end{figure*}

\section{Approach}

\subsection{Overall Architecture}
The overall pipeline of HybridOcc is shown in Fig. \ref{Fig.Overall pipline}. Taking the multi-camera images as the inputs, we use an image backbone to extract multi-scale camera features. Then we learn the sparse 3D volume features through a dual branch composed of Transformer framework and NeRF representation. Specifically, the Transformer branch learns 3D volume-shaped queries from multi-camera features via a 2D to 3D transformation module. The hybrid 3D query proposals are derived from Transformer and NeRF respectively, and gradually refined in a coarse-to-fine manner (see Sec. \ref{sec:3.3}). In the NeRF branch, the vanilla NeRF paradigm is replaced by the new autonomous driving scene occupancy prediction NeRF module. Volume rendering occupancy prediction models are directly supervised by depth rather than RGB color (see Sec. \ref{sec:3.4}). The semantic occupancy ground truth supervises multi-scale volume semantic occupancy prediction.

\subsection{Transformer Branch}
\label{sec:3.3}

\textbf{Coarse-to-fine Approach.} Unlike the dense 3D volume obtained in SurroundOcc \cite{wei2023surroundocc}, inspired by Occ3D \cite{tian2024occ3d}, we adopt a coarse-to-fine approach to gradually refine the sparse volume, as shown in the upper part of Fig. \ref{Fig.Overall pipline}. Specifically, the semantic occupancy $O_{l}$ of the 3D volume space $V_{l}\in \mathbb{R}^{{H_{l}\times W_{l}\times Z_{l}\times C}}$ at each scale is predicted by an MLP following SurroundOcc \cite{wei2023surroundocc}. Voxels with an occupancy value lower than the occupancy threshold $\theta$ are defined as empty voxels. The $l$-th volume occupancy $O_{l}$ serves as part of the query prior position distribution in the higher resolution of the volume $V_{l+1}$, as shown in the purple arrow and purple square in Fig. \ref{Fig.Overall pipline}. Sparse voxels of $V_{l+1}$ are recorded as sparse query proposals $Q_{l+1,s}\in \mathbb{R}^{N_{l+1}\times C}, Q_{l+1,s}\subseteq V_{l+1}$, and $Q_{l+1,s}$ is learned from multi-camera features at each scale via the 2D to 3D module. Finally, $Q_{l+1,s}$ is skip-connected with upsampled $Q_{l,s}$ and fed to the MLP to predict $l$+1-th layer semantic occupancy. The semantic occupancy prediction can be expressed as:
\begin{equation}
    O_{l+1}=h\left(Q_{l+1,s}+\operatorname{up}\left(Q_{l,s}\right)\right),
\end{equation}

\noindent where up is 2$\times$ upsample, $h(\cdot)$ represent MLP. 

It is worth noting that the initial query proposals of the coarse-grained volume $V_{1}$ are densely constructed. The prior spatial distribution of query proposals of the fine-grained $V_{2}$, $V_{3}$, and $V_{4}$ are composed of the hybrid of binary occupancies of the Transformer branch and the NeRF branch at each scale, respectively (see Sec. \ref{sec:3.4}).

\textbf{2D to 3D Transformation.} Inspired by recent Transformer-based multi-camera 3D perception methods \cite{wei2023surroundocc,li2022bevformer}, we project the 3D reference points of the volume onto the 2D camera to aggregate features. Specifically, each 3D reference point corresponding to query $q\subseteq Q_{l}$, is projected to the 2D feature map according to the given camera intrinsic and extrinsic parameters and performs deformable cross-attention (DeformAtt) to learn features:
\begin{equation}
\operatorname{DeformAtt}\left(q,X\right)=\sum_{m=1}^{M} W_{m} \sum_{k=1}^{K} A_{mk} \cdot W_{m}^{\prime}X\left(p+\Delta p_{mk}\right),
\end{equation}

\noindent where $X$ is the multi-camera features, $W_{m}$ and $W_{m}^{\prime}$ are the weights obtained by linear projection, $A_{mk}$ is the attention weights and $A_{mk}\in [0, 1]$, $X\left(p+\Delta p_{mk}\right)$ are the sampled features corresponding to the 2D reference point $p$, $\Delta p_{mk}$ is the learned position offsets corresponding to $p$. Other settings follow SurroundOcc \cite{wei2023surroundocc} and BEVFormer \cite{li2022bevformer}. Finally, the volume-shaped query $Q$ is further optimized through 3D sparse convolution, so that each voxel query subset pays attention to local information of each other.

\subsection{Neural Radiance Field Branch}
\label{sec:3.4}

\textbf{Depth Rendering Supervision.} Vanilla NeRF \cite{mildenhall2021nerf,hu2022efficientnerf} optimizes a continuous radiance field $f\left(\cdot \right)=\left(c,\rho \right)$ based on the density $\rho$ of sample points along the ray, and supervises volume rendering with RGB. The difference is that we design a new radiance field based on SceneRF \cite{cao2023scenerf}, and the new NeRF model has depth supervision to predict 3D occupancy. 

The NeRF branch is shown at the bottom of Fig. \ref{Fig.Overall pipline}. The NeRF branch performs occupancy prediction and depth rendering based on $l$-th level multi-camera features $X_{l}, l=2,3,$ and $4$ from the image backbone. We uniformly sampled \textit{I} pixels from the pixel coordinates of each camera, and sample \textit{N} points along each ray passing through these pixels. The uniform sampling strategy is consistent with SceneRF \cite{cao2023scenerf}. 
Then following SceneRF converts $X_{l}$ into a spherical space to obtain $X_{i}^{\prime}$, so that each sampled point \textit{x} can be projected on the spherical space for retrieving the image feature vector $X_{l}^{\prime}(x)$ through bilinear interpolation. Finally, the feature $X_{l}^{\prime}(x)$ of the point \textit{x} and the 3D position encoding $\gamma\left(x\right)$ are fed to the implicit expression function MLP to predict the binary occupancy $\sigma_{l}$ of volume $V_{l}$. Note that the NeRF branch only needs to provide prior spatial distribution information of query for the Transformer-based coarse-to-fine structure, therefore, we only predict class-agnostic occupancy. The binary occupancy prediction implicit radiance field is defined as:

\begin{equation}
   f\left(\gamma\left(x\right), X_{l}^{\prime}(x)\right)=\left(d_{l},\sigma_{l} \right), 
\end{equation}

\noindent where $d_{l}$ is the $l$-th scale depth.

Unlike most NeRFs \cite{mildenhall2021nerf,yu2021pixelnerf} that use volume rendering from density to predict colors, we attempt to reveal depth explicitly from the radiance volume. Depth volume rendering is performed on multi-scale image features $X_{2}$, $X_{3}$, and $X_{4}$ respectively, so that multi-scale features can obtain depth supervision. For $l$-th scale feature, we define depth volume rendering as:
\begin{equation}
\begin{split}
\textit{D}_{\textit{r},l}&=\sum_{i=1}^{N}w_{i,l}d_{i,l}, \\
w_{i,l} = {T}_{i,l}&\left(1-exp(-\sigma_{i,l} \delta_{i,l})\right),
\end{split}  
\end{equation}
where $\textit{D}_{\textit{r},l}$ is $l$-th ray depth, $T_{i,l}$ is the accumulated transmittance, $T_{i,l}=-\sum_{i=1}^{n-1}\sigma_{i,l} \delta_{i,l}$, and $\delta_{i,l}$ is the distance to the previous adjacent point.

\begin{figure}[htb]
  \centering
   \centerline{\includegraphics[width=1.0\linewidth]{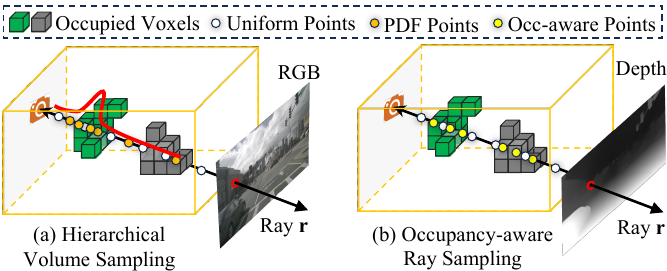}}
   \caption{Comparisons of the proposed occupancy-aware ray sampling with hierarchical volume sampling. Our ray sampling strategy focuses on sampling all visible or invisible occupied voxels passing through the ray.}
   \label{fig3}
\end{figure}

\textbf{Occupancy-aware Ray Sampling.} Previous research \cite{mildenhall2021nerf,zhang2018efficient,cao2023scenerf} has shown that sampling points along the ray near the surface of the scene can effectively increase rendering efficiency. Fig. \ref{fig3}(a) illustrates the hierarchical volume sampling \cite{mildenhall2021nerf} method produces a probability density function (PDF) focused on the surface along the ray to optimize the sampling points. In the NeRF module of the SSC task, the objective is to estimate the depth based on features of visible surfaces and invisible voxels, rather than the color or category. The implicit function of the NeRF branch needs to complete the 3D volume binary occupancy prediction of the entire scene. As for the occupancy prediction NeRF model, intuitively, the sampling points along the ray falling on non-empty voxels can improve the rendering. Therefore, we propose an occupancy-aware ray sampling strategy in which the occupancy prediction result $O_{l}$ of volume $V_{l}$ explicitly guides each sampling point along the ray in the volume $V_{l+1}$, as shown in Fig. \ref{fig3}(b).

Specifically, for each ray, we first sample 128 points uniformly between the near and far boundaries. These points are then projected into 3D volume $V_{3}$ to query the occupancy state $O_{3}$, and we sample 32 points based on the occupancy state. If more than 32 points are occupied, 32 points are randomly sampled from them. Otherwise, we accept all occupied points and randomly sample the remaining points. The occupancy-aware ray sampling strategy concentrates on sampling occupied voxels within the scene, optimizing binary occupancy prediction and volume rendering of depth.

\textbf{Hybrid Query Proposals.} The aforementioned NeRF branch has the 3D occupancy prediction capability. Take the nuScenes \cite{caesar2020nuscenes} dataset as an example, we first independently partition each camera's features $X_{2}$, $X_{3}$, and $X_{4}$ into the 3D voxel space, and predict the binary occupation of the 3D volume by the NeRF module. Then, we fuse the multi-camera results in volume coordinate $V_{\operatorname{imp},l}^{H_{l}\times W_{l}\times Z_{l}}$ with camera extrinsics and obtain the occupancy distribution $O_{\operatorname{imp},l}\in \left\{0; 1\right\}^{H_{l}\times W_{l}\times Z_{l}}$. Meanwhile, the depth supervision signal also updates the image features, making the model sensitive to depth. Finally, we fuse the implicitly predicted occupancy $O_{\operatorname{imp},l}$ from the NeRF branch with the explicitly estimated $O_{\operatorname{exp},l}$ from the coarse-grained Transformer branch in volume $V_{l-1}$. The hybrid query proposals serve as the $l$-th level query to attend to the 2D to 3D process. Hybrid query proposals $Q_{l,s}$ can be expressed as:

\begin{equation}
   Q_{l,s}=Q\left[O_{\operatorname{imp},l}\right] \cup Q\left[O_{\operatorname{exp},l}\right].
\end{equation}

\subsection{Loss Fusion}
Our dual-branch occupancy prediction network is an end-to-end optimization model. The overall loss of the model $L_{total} =L_{\operatorname{exp}}+\beta L_{\operatorname{imp}}$, where $L_{\operatorname{exp}}$ is the explicit loss of the Transformer branch, $L_{\operatorname{imp}}$ is the implicit loss of the NeRF branch, and $\beta$ is set to 0.5. 
The supervision of multi-scale 3D volumes is inspired by SurroundOcc \cite{wei2023surroundocc}. We also supervise each scale volume to get coarse-grained and fine-grained 3D features. We adopt cross-entropy loss for 3D semantic occupancy prediction. $L_{\operatorname{exp}}$ is expressed as:
\begin{equation}
   L_{\operatorname{exp}}=\sum_{i=1}^{L} \alpha_{i}L_{i}\left(\sigma_{i}\right),
\end{equation}
\noindent where $\alpha_{i}$ is the decayed loss weight for the \textit{l}-th scale supervision. Hybrid queries of high-resolution volume $V_{4}$ require sufficient supervision signals. 

For the loss of the NeRF branch, we use binary cross-entropy loss for class-agnostic occupancy prediction and utilize SILog loss \cite{Eigen2014DepthMP} to optimize the depth, the depth is supervised by the projection of the LiDAR points. Also includes the decayed loss weight $\alpha_{i}$. $L_{\operatorname{imp}}$ is expressed as:
\begin{equation}
   L_{\operatorname{imp}}=\sum_{i=2}^{L} \alpha_{i}(L_{\operatorname{depth},i}(\textit{D}_{\textit{r}},\widehat{\textit{D}}_{\textit{r}})+L_{i}\left(\sigma\right)).
\end{equation}

\section{EXPERIMENT}
\subsection{Datasets}

We conduct multi-camera semantic scene completion experiments on the nuScenes dataset \cite{caesar2020nuscenes}, which contains surround RGB image data from 6 cameras and Lidar sweeps covering the full 360-degree field of view. The 1 000 multi-modal data are split into train/val/test splits with 700/150/150. SurroundOcc \cite{wei2023surroundocc} proposed a 3D SSC benchmark based on the nuScenes dataset, and there are 17 categories of 3D occupancy. The perception range is clipped into $[$-$50m, 50m]$ for \textit{X}, \textit{Y} axis and $[$-$5m, 3m]$ for \textit{Z} axis. The ground truth volume dimension of semantic occupancy is 200$\times$200$\times$16 with $0.5m$ voxel size. For Occ3D-nuScenes, The perception range is clipped into $[$-$40m, 40m]$ for \textit{X}, \textit{Y} axis and $[$-$1m, 5,4m]$ for \textit{Z} axis. The final output occupancy shape is 200$\times$200$\times$16 with $0.4m$ voxel size.

To further demonstrate the effectiveness of our method, we conduct monocular semantic scene completion experiment on the SemanticKITTI dataset \cite{behley2019semantickitti}, which annotates autonomous driving scene with 21 semantic classes (19 semantics, 1 free, and 1 unknown). The dataset contains 22 sequences and is split into 10/1/11 for train/val/test. The perception range is clipped into $[$-$25.6m, 25.6m]$ for \textit{X} axis, $[0, 51.2m]$ for \textit{Y} axis, and $[-2m, 4.4m]$ for \textit{Z} axis. The ground truth semantic occupancy has a dimension of 256$\times$256$\times$32 with $0.2m$ voxel size.

\subsection{Metrics}
For both SurroundOcc-nuScenes \cite{wei2023surroundocc} and SemanticKITTI \cite{behley2019semantickitti} datasets, we report the intersection over union (IoU) of occupied voxels as the evaluation metric of the class-agnostic scene completion (SC) task and the mIoU of all semantic classes for the SSC task following SurroundOcc \cite{wei2023surroundocc}. For Occ3D-nuScenes \cite{tian2024occ3d}, we report the mIoU following FB-Occ \cite{li2023fb} and Occ3D \cite{tian2024occ3d}. It is worth noting that Occ3D-nuScenes only evaluates visible regions, as shown in Fig. \ref{figoccupancy prediction_compa}. We refer the readers to previous papers\cite{wei2023surroundocc,behley2019semantickitti,tian2024occ3d} for more details.


\subsection{Implementation Details}
For the SurroundOcc-nuScenes \cite{wei2023surroundocc,caesar2020nuscenes}dataset, the input image resolution is 900$\times$1600. We following SurroundOcc \cite{wei2023surroundocc} adopt ResNet-101 \cite{res,Dai2017DeformableCN} initialized from the FCOS3D \cite{wang2021fcos3d} checkpoint as the image backbone. The image backbone yields 3 level feature maps, and employs FPN \cite{fpn} following the backbone to produce 4 level feature maps with hidden dimensions 256. For the SemanticKITTI dataset, we crop image of cam2 to size 370$\times$1220, and following \cite{wei2023surroundocc,Zhang2023OccFormerDT} use EfficientNetB7 \cite{cao2022monoscene} as the image backbone for fair comparison. For the Occ3D-nuScenes \cite{wei2023surroundocc,caesar2020nuscenes}dataset, we follow FB-Occ \cite{li2023fb} resize the input image resolution to 256$\times$704 and adopt ResNet-50 as image backbone.

For both SurroundOcc-nuScenes and SemanticKITTI datasets, we set the number of 2D to 3D spatial cross-attention layers as 6, 3, 1, and 3 respectively. Each level of spatial cross-attention uses 8, 4, 4, and 4 sampling points around each reference point, respectively. The MLP structure of the implicit function $f\left(\cdot \right)$ in NeRF module is consistent with that of SceneRF \cite{cao2023scenerf}. We use 32 points per ray in occupancy-aware ray sampling. The occupancy threshold $\theta$ is set to 0.5. The Occ3D-nuScenes dataset evaluates the visible surfaces of a scene, and we simply set our HybridOcc sampling strategy to probabilistic ray sampling \cite{cao2023scenerf}. We train our model 24 epochs on nuScenes dataset and 30 epochs on SemanticKITTI dataset with a learning rate $2\times10^{-4}$ by default. All models are trained with a batch size of 4 on 4 NVIDIA A800 GPUs.

\begin{table*}[htb]
\renewcommand\arraystretch{1.15}
\centering
\caption{Semantic scene completion results on SurroundOcc-nuScenes val set. * indicates SurroundOcc \cite{wei2023surroundocc} reports these models. $\dagger$ represents trained on SurroundOcc-nuScenes. FB-Occ(D) \cite{li2023fb} is the single-frame version.}
\tabcolsep=0.1cm
\footnotesize
\begin{tabular}{l|cc|cccccccccccccccc}

\hline
Method       & \begin{tabular}[c]{@{}l@{}} IoU\end{tabular} & \begin{tabular}[c]{@{}l@{}} mIoU\end{tabular}  & \rotatebox{90}{barrier} & \rotatebox{90}{bicycle} & \rotatebox{90}{bus}   & \rotatebox{90}{car}   & \rotatebox{90}{const. veh.} & \rotatebox{90}{motorcycle} & \rotatebox{90}{pedestrian} & \rotatebox{90}{traffic cone} & \rotatebox{90}{trailer} & \rotatebox{90}{truck} & \rotatebox{90}{drive. suf.} & \rotatebox{90}{other flat} & \rotatebox{90}{sidewalk} & \rotatebox{90}{terrain} & \rotatebox{90}{manmade} & \rotatebox{90}{vegetation} \\
\hline
BEVFormer* \cite{li2022bevformer}   & 30.50   & 16.75   & 14.22   & 6.58  & 23.46 & 28.28 & 8.66  & 10.77  & 6.64   & 4.05   & 11.20  & 17.78 & 37.28   & 18.00   & 22.88  & 22.17   & 13.80  & 22.21  \\
TPVFormer* \cite{huang2023tri}  & 30.86  & 17.10   & 15.96   & 5.31  & 23.86 & 27.32 & 9.79  & 8.74  & 7.09  & 5.20  & 10.97  & 19.22 & 38.87  & 21.25  & 24.26  & 23.15   & 11.73   & 20.81   \\
FB-Occ(D)$\dagger$ \cite{li2023fb} &31.56  &20.17 &20.31 &12.29 &26.33 &31.07 &10.78 &15.95 &13.31 &11.14 &13.24 &22.13 &39.56 &22.26 &25.14 &23.59 &13.92 &21.64      \\

SurroundOcc* \cite{wei2023surroundocc}  & 31.49  & 20.30  & 20.59   & 11.68   & 28.06 & 30.86 & 10.70  & 15.14  & 14.09  & 12.06  & 14.38   & 22.26 & 37.29   & 23.70   & 24.49  & 22.77   & 14.89   & 21.86  \\
OccFormer$\dagger$ \cite{Zhang2023OccFormerDT} &31.54  &20.97 &21.98 &11.92 &28.77 &31.91 &\textbf{11.62} &14.92 &\textbf{14.26} &11.57 &\textbf{15.38} &23.60 &40.01 &22.93 &25.74 &24.14 &14.51 &22.29      \\
\hline
\textbf{HybridOcc(Ours)} &\textbf{33.07}  &\textbf{21.36}  &\textbf{22.29}   &12.13   &\textbf{29.78}   &\textbf{32.34}   &10.94   &\textbf{16.33}   &14.07  &\textbf{12.69}  &14.63   &\textbf{23.98}   &\textbf{40.43}   &\textbf{23.69}  &\textbf{26.15}  &\textbf{24.53}   &\textbf{15.23}   &\textbf{22.60}   \\  
\hline

\end{tabular}
\label{tab:table1}
\end{table*}

\begin{table*}[htb]
\renewcommand\arraystretch{1.15}
\centering
\caption{Monocular semantic scene completion results on SemanticKITTI test set. Our method surpasses VoxFormer \cite{li2023voxformer}, which uses an additional depth prediction network \cite{shamsafar2022mobilestereonet}.}
\tabcolsep=0.1cm
\footnotesize
\begin{tabular}{l|c|c|ccccccccccccccccccc}
\hline
Method  & \begin{tabular}[c]{@{}l@{}} IoU \end{tabular} & \begin{tabular}[c]{@{}l@{}}mIoU\end{tabular} & \rotatebox{90}{\begin{tabular}[c]{@{}l@{}}road \end{tabular}} & \rotatebox{90}{\begin{tabular}[c]{@{}l@{}}sidewalk \end{tabular}} & \rotatebox{90}{\begin{tabular}[c]{@{}l@{}}parking \end{tabular}} & \rotatebox{90}{\begin{tabular}[c]{@{}l@{}}other-grnd \end{tabular}} & \rotatebox{90}{\begin{tabular}[c]{@{}l@{}}building\end{tabular}} & \rotatebox{90}{\begin{tabular}[c]{@{}l@{}}car\end{tabular}} & \rotatebox{90}{\begin{tabular}[c]{@{}l@{}}truck\end{tabular}} & \rotatebox{90}{\begin{tabular}[c]{@{}l@{}}bicycle\end{tabular}} & \rotatebox{90}{\begin{tabular}[c]{@{}l@{}}motorcycle\end{tabular}} & \rotatebox{90}{\begin{tabular}[c]{@{}l@{}}other-veh. \end{tabular}} & \rotatebox{90}{\begin{tabular}[c]{@{}l@{}}vegetation\end{tabular}} & \rotatebox{90}{\begin{tabular}[c]{@{}l@{}}trunk\end{tabular}} & \rotatebox{90}{\begin{tabular}[c]{@{}l@{}}terrain\end{tabular}} & \rotatebox{90}{\begin{tabular}[c]{@{}l@{}}person\end{tabular}} & \rotatebox{90}{\begin{tabular}[c]{@{}l@{}}bicyclist\end{tabular}} & \rotatebox{90}{\begin{tabular}[c]{@{}l@{}}motorcyclist.\end{tabular}} & \rotatebox{90}{\begin{tabular}[c]{@{}l@{}}fence\end{tabular}} & \rotatebox{90}{\begin{tabular}[c]{@{}l@{}}pole \end{tabular}} & \rotatebox{90}{\begin{tabular}[c]{@{}l@{}}traf.-sign \end{tabular}} \\
\hline
MonoScene \cite{cao2022monoscene}  & 34.16  & 11.08   & 54.70  & 27.10  & 24.80   & 5.70  & 14.40  & 18.80  & 3.30  & 0.50  & 0.70  & 4.40  & 14.90   & 2.40  & 19.50   & 1.00  & 1.40   & \textbf{0.40}  & 11.10  & 3.30  & 2.10  \\
TPVFormer \cite{huang2023tri}  & 34.25  & 11.26   & 55.10  & 27.20  & 27.40   & 6.50  & 14.80  & 19.20  & \textbf{3.70}  & 1.00   & 0.50  & 2.30  & 13.90   & 2.60  & 20.40   & 1.10   & \textbf{2.40}   & 0.30  & 11.00  & 2.90  & 1.50  \\
SurroundOcc \cite{wei2023surroundocc}  & 34.72  & 11.86   & 56.90  & 28.30  & 30.20   & 6.80  & 15.20  & 20.60  & 1.40  & 1.60  & 1.20  & 4.40  & 14.90   & 3.40  & 19.30   & 1.40  & 2.00  & 0.10  & 11.30  & 3.90  & 2.40  \\
OccFormer \cite{Zhang2023OccFormerDT} & 34.53  & 12.32   & 55.90  & 30.30  & \textbf{31.50}   & 6.50  & 15.70  & 21.60  & 1.20  & 1.50  & 1.70  & 3.20  & \textbf{16.80}   & 3.90  & 21.30   & \textbf{2.20}   & 1.10   & 0.20  & 11.90  & 3.80  & 3.70  \\

VoxFormer \cite{li2023voxformer} & \textbf{44.02}  & 12.35   & 54.76  & 26.35  & 15.50   & 0.70  & 17.65  & 25.79  & 5.08  & 0.59  & 0.51  & 3.77  & 24.39   & 5.63  & 29.96   & 1.78   & 3.32   & 0.00  & 7.64  & 7.11  & 4.18  \\

\hline
HybridOCC(Ours)  &36.34  &\textbf{12.68}   &\textbf{57.38}   &\textbf{31.10}  &30.92  &\textbf{7.20}  &\textbf{16.20}  &\textbf{21.70}  &1.60   &\textbf{1.70}  &\textbf{1.80}   &\textbf{4.70}  &16.20   &\textbf{4.20}   &\textbf{21.50}  &2.10  &2.20  &0.20   &\textbf{12.20}  &\textbf{4.20}  &\textbf{3.90}   \\
\hline
\end{tabular}
\label{tab:table2}
\end{table*}

\begin{table}[htb]
\renewcommand\arraystretch{1.15}
\centering
\caption{3D occupancy prediction performance on Occ3D-nuScenes val set. * means reported by CTF-Occ \cite{tian2024occ3d}, $\dagger$ means only 2D Supervision. FB-Occ(D) \cite{li2023fb} is the single-frame version.}
\tabcolsep=0.2cm
\begin{tabular}{l|c|c|c}
\hline
Method      & Backbone   & Revolution   & mIoU  \\
\hline
SelfOcc$\dagger$ \cite{huang2023selfocc}     & ResNet-50 & 768$\times$1600     & 9.30 \\
OccNeRF$\dagger$ \cite{zhang2023occnerf}     & ResNet-101 & 900$\times$1600     & 10.81 \\
OccFormer* \cite{Zhang2023OccFormerDT} & ResNet-101 & 900 $\times$ 1600 & 21.93 \\
RenderOcc \cite{pan2023renderocc}  & ResNet-101 & 512$\times$1408     & 23.95 \\
BEVFormer* \cite{li2022bevformer} & ResNet-101  & 900 $\times$ 1600 & 26.88 \\
TPVFormer* \cite{huang2023tri}  & ResNet-101  & 900 $\times$ 1600 & 27.83 \\
CTF-Occ* \cite{tian2024occ3d}    & ResNet-101  & 900 $\times$ 1600 & 28.53 \\
SurroundOcc \cite{wei2023surroundocc} & ResNet-50   & 256 $\times$ 704   & 36.32 \\
FB-Occ(D) \cite{li2023fb}     & ResNet-50   & 256 $\times$ 704   & 37.39 \\
\hline
HybridOcc   & ResNet-50   & 256 $\times$ 704   & \textbf{37.82} \\
\hline
\end{tabular}
\label{tab:occ3dresult}
\end{table}

\subsection{Main Results}
In Table \ref{tab:table1}, we report the multi-camera semantic scene completion task on SurroundOcc-nuScenes \cite{wei2023surroundocc} val set. We compare our HybridOcc with several vision-based methods \cite{wei2023surroundocc,Zhang2023OccFormerDT,li2023fb}. Compared with end-to-end OccFormer \cite{Zhang2023OccFormerDT}, HybridOcc achieves a 1.53\% IoU lead. OccFormer has a learning-based deep prediction network and contains two Transformer modules for aggregating contextual features. For the Transformer-based multi-scale SurroundOcc \cite{wei2023surroundocc}, our hybrid query strategy brings a remarkable boost of 1.58\% IoU and 1.06\% mIoU. Compared with the single-frame variant of FB-Occ \cite{li2023fb}, HybridOcc achieves 1.51\% IoU and 1.19\% mIoU lead. FB-Occ has a pre-trained depth prediction network, but lacks consideration of occluded invisible voxels. Our HybridOcc can achieve end-to-end training while the NeRF branch additionally considers invisible voxels.

In addition, we conduct semantic scene completion experiments on the monocular SemanticKITTI dataset, as shown in Table \ref{tab:table2}. HybridOcc outperforms several existing competitors \cite{wei2023surroundocc,Zhang2023OccFormerDT,li2023voxformer}. It achieves approximately 12.68\% mIoU on the SSC task, which achieves an improvement of 0.33\% compared to two-stage VoxFormer\cite{li2023voxformer}. VoxFormer uses an additional depth prediction network \cite{shamsafar2022mobilestereonet} to provide strong position priors for 3D voxel query proposals, achieving a remarkable 44.02\% IoU. However, VoxFormer's ablation experiments show that without adding a one-stage depth prediction network, it only has 34.64\% IoU. Compared to this variant, our HybridOcc achieves a 1.7\% IoU lead. Compared with the end-to-end OccFormer \cite{Zhang2023OccFormerDT}, which also has a depth prediction network, HybridOcc shows a 1.81\% IoU lead. Monocular SSC experiments also demonstrate the effectiveness of our method.

In Table \ref{tab:occ3dresult}, we further experiment on Occ3D-nuScenes \cite{tian2024occ3d}, which only evaluates visible voxels. HybridOcc achieves 37.82\% mIoU, surpassing FB-Occ \cite{li2023fb} with the depth network by 0.43\% mIoU. Furthermore, HybridOcc achieves 1.50\% mIoU lead compared to SurroundOcc \cite{wei2023surroundocc}, which uses an attention-based dense multi-scale supervision framework.

\begin{figure*}[t]
\centering
\centerline{\includegraphics[width=1\textwidth]{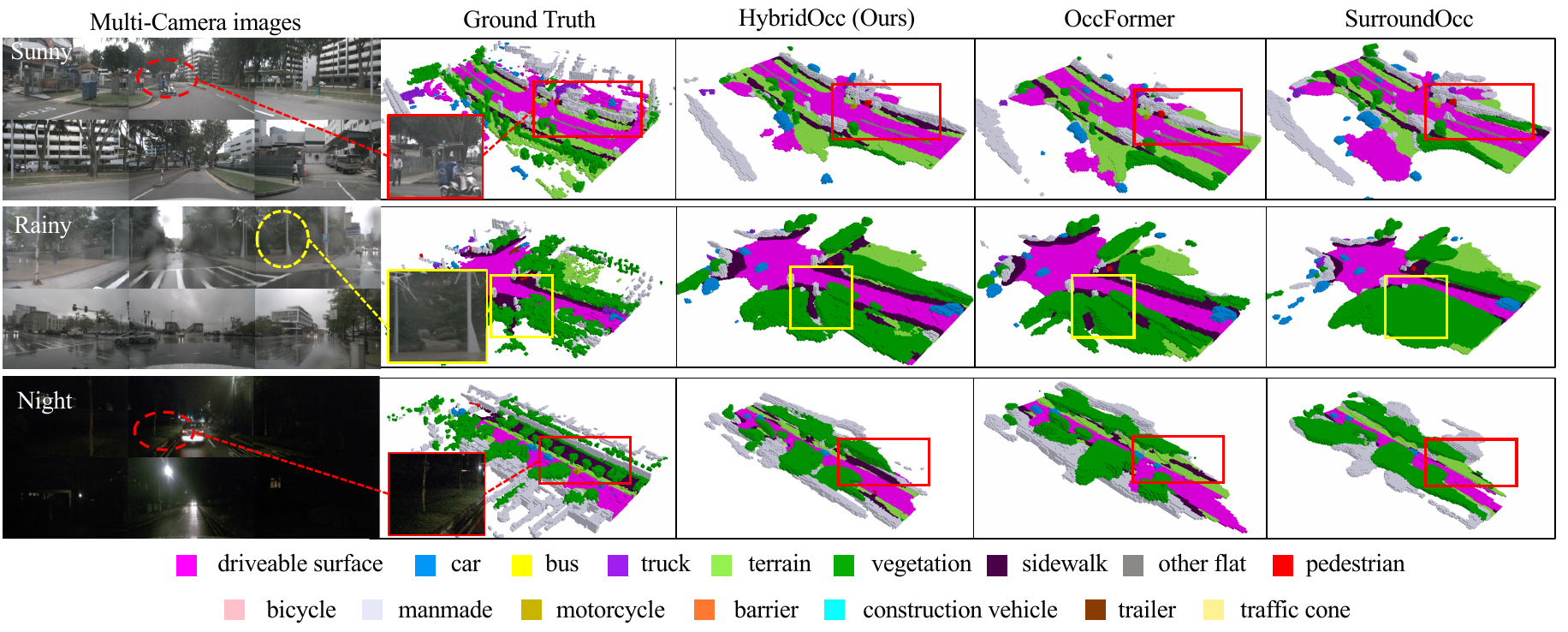}}
\caption{Visualizations on nuScenes validation set. The leftmost column is the input multi-camera surround image, and the next four columns show the ground truth semantic occupancy, the 3D semantic occupancy predicted by our HybridOcc, OccFormer \cite{Zhang2023OccFormerDT} and SurroundOcc \cite{wei2023surroundocc}.}
\label{Fig.Visualizations}
\end{figure*}

\begin{table}[]
\renewcommand\arraystretch{1.15}
\centering
\caption{Ablation study for architectural components. w.o. denotes without.}
\setlength{\tabcolsep}{2.4mm}{
\footnotesize
\begin{tabular}{@{}l|c|cc@{}}
\hline
Method                                        & SC IoU & SSC mIoU \\
\hline
w.o. Transformer branch                  &  28.84      & 15.49         \\
w.o. NeRF branch                         &  32.19      & 20.61         \\
w.o. Sparse convolution                 &  32.87      & 21.16         \\
Ours                                    &  \textbf{33.07}  &\textbf{21.36}         \\
\hline
\end{tabular}
}
\label{tab:architectural components}
\end{table}

\subsection{Ablation Studies} 
\label{sec:Ablation Studies}

\textbf{Architectural components.} In Table \ref{tab:architectural components}, we ablate the architectural components in our HybridOcc. Since our proposed NeRF branch cannot operate independently of the Transformer structure, we use the probabilistic ray sampling method \cite{cao2023scenerf} in the NeRF model to ablate the case without the Transformer branch. The NeRF model shows certain performance on the SSC task. The $1^{st}$ and $2^{nd}$ rows indicate that the NeRF-based method performs worse than the Transformer-based method. In $3^{rd}$ row, 3D sparse convolution positively contributes to performance. Finally, Hybrid query proposals include depth and invisible information supplementation from the NeRF branch, which improves the occupancy prediction precision.

\textbf{Ablation on Ray Sampling.} In Table \ref{tab:Ray Sampling}, we conduct a comprehensive study on the ray sampling strategy of the neural radiance field branch. Probabilistic ray sampling \cite{cao2023scenerf} concentrates sampling points near the geometric surface. This type of approach performs worse than occupancy-aware ray sampling. The potential reason is that voxel features of visible surfaces and occluded regions are needed to jointly optimize the radiance field in the SC task. Additionally, due to the sparsity of the scene, 32 sampling points per ray are sufficient to optimize a single view with a sensing range of $50m$.

\begin{table}[]
\renewcommand\arraystretch{1.15}
\centering
\caption{Ablation study for ray sampling, we vary point number and sampling method.}
\footnotesize
\begin{tabular}{@{}l|c|cc@{}}
\hline
Method                                                                            & \begin{tabular}[c]{@{}l@{}}Point\end{tabular} & \begin{tabular}[c]{@{}l@{}}SC IoU\end{tabular} &\begin{tabular}[c]{@{}l@{}}SSC mIoU\end{tabular} \\
\hline
\begin{tabular}[c]{@{}l@{}}Hierarchical volume sampling \cite{mildenhall2021nerf} \end{tabular}            & 64         & 32.51       &20.79          \\
\begin{tabular}[c]{@{}l@{}}Probabilistic ray sampling \cite{cao2023scenerf} \end{tabular}              & 32          & 32.70       & 20.82         \\
\hline
\multirow{3}{*}{\begin{tabular}[c]{@{}l@{}}Occupancy-aware ray sampling\end{tabular}} & 16      & 32.56      & 20.85         \\
                                                                                  & 32          & 33.07       &\textbf{21.36}        \\
                                                                                  & 64          & \textbf{33.10}       & 21.28         \\
\hline
\end{tabular}
\label{tab:Ray Sampling}
\end{table}

\begin{table}[]
\renewcommand\arraystretch{1.15}
\caption{The model efficiency and performance comparison on SurroundOcc-nuScenes \cite{wei2023surroundocc} dataset. The latency of all methods is measured on a single A800 GPU.}
\begin{tabular}{l|cccc}
\hline
Method                     & mIoU  & Memory(G) & Latency(ms) \\
\hline
FB-Occ(D) \cite{li2023fb}      &20.17   & \textbf{5.40}         & \textbf{337}        \\
SurroundOcc \cite{wei2023surroundocc}           &20.30  & 5.90     & 413   \\
\hline
HybridOcc(NeRF\cite{cao2023scenerf})       & \textbf{21.36}  & 7.40      & 522            \\
HybridOcc(Instant-NGP\cite{muller2022instant}) & 21.29  &  6.30      & 426            \\
\hline 
\end{tabular}
\label{tab:efficiency}
\end{table}

\textbf{Ablation on efficiency of NeRF branch.} In Table \ref{tab:efficiency}, we study the impact of the NeRF variant on model efficiency and compare it with SurroundOcc \cite{wei2023surroundocc} and FB-Occ \cite{li2023fb}. There are two potential possibilities for accelerating NeRF on the SSC task. One is to reduce sampling points, and the other is to reduce the number and dimensions of MLP layers in NeRF. We adopt the lightweight MLP proposed by Instant-NGP \cite{muller2022instant} to accelerate our model. The results show that Instant-NGP reduces the inference memory usage and the inference latency (Instant-NGP variant (55 ms) $v.s.$ NeRF (151 ms)). The introduction of the NeRF branch inevitably makes its inference speed slower than FB-Occ, but our model does not require an additional pre-trained depth network and has slightly higher performance.

\subsection{Visualization}

Fig. \ref{Fig.Visualizations} shows a visualization of the SSC prediction results of proposed HybridOcc, OccFormer \cite{Zhang2023OccFormerDT} and SurroundOcc \cite{wei2023surroundocc} on the SurroundOcc-nuScenes \cite{wei2023surroundocc} dataset. We show sunny day, rainy day, and night visualization. Compared to OccFormer and SurroundOcc, which have large and continuous occupancies, our HybridOcc results are much more refined. Such as objects of different categories have more independent and complete boundaries, which also reduces the predicted false positive phenomenon (yellow marks). In addition, HybridOcc shows better semantic refinement capabilities, such as better distinction between terrain and sidewalk (red marks) that are approximately the same plane.

\section{CONCLUSION}
In this paper, we have presented HybridOcc, which generates sparse hybrid 3D volume query proposals by Transformer and NeRF for semantic scene completion prediction. To more precisely lift the 2D camera features to 3D volumes, the proposed hybrid query from coarse to fine gradually refines autonomous driving scenes. The newly designed NeRF branch implicitly infers the visible and invisible scene occupancy, explicitly obtains depth signal supervision, and combines it with the Transformer branch to hallucinate the overall volume semantic layout of the scene with higher accuracy. Considering that the occupancy prediction NeRF branch relies on non-empty voxel features, we have designed an occupancy-aware ray sampling method to optimize the hybrid model significantly. The superiority of our HybridOcc is demonstrated by comparing the nuScenes and SemanticKITTI datasets.


\clearpage

\bibliographystyle{IEEEtran}
\bibliography{refs}

\vfill

\end{document}